\documentclass[twocolumn,10pt]{asme2e}
\usepackage[]{graphicx, epstopdf}
\usepackage{amsmath} 
\usepackage{amsfonts}
\usepackage{amssymb}
\usepackage{bm}
\usepackage{algorithm, algorithmicx}
\usepackage[]{algpseudocode}
\usepackage{color}
\usepackage{subcaption}
\usepackage{caption}
\usepackage{comment}
\usepackage[dvipsnames]{xcolor}
\usepackage[left]{lineno}
\usepackage{float}

\newtheorem{remark}{Remark} 	

\graphicspath{ {figures/} }

\confshortname{IDETC/MR 2019}
\conffullname{the ASME 2019 International Design Engineering Technical Conferences \&\\
              Computers and Information in Engineering Conference}

\confdate{18-21}
\confmonth{August}
\confyear{2019}
\confcity{Anaheim, CA}
\confcountry{USA}

\papernum{DETC2019/MR-97945}

\title{Computing Robust Inverse Kinematics Under Uncertainty}

\author{Anirban Sinha
    \affiliation{
	Department of Mechanical Engineering\\
	Stony Brook University, New York, USA\\
    Email: anirban.sinha@stonybrook.edu
    }	
}

\author{Nilanjan Chakraborty
    \affiliation{
	Department of Mechanical Engineering\\
	Stony Brook University, New York, USA\\
    Email: nilanjan.chakraborty@stonybrook.edu
    }	
}

\begin{document}
\maketitle    
\begin{abstract}
\noindent
{\it Robotic tasks, like reaching a pre-grasp configuration, are specified in the end effector space or task space, whereas, robot motion is controlled in joint space. Because of inherent actuation errors in joint space, robots cannot achieve desired configurations in task space exactly. Furthermore, different inverse kinematics (IK) solutions map joint space error set to task space differently. Thus for a given task with a prescribed error tolerance, all IK solutions will not be guaranteed to successfully execute the task. Any IK solution that is guaranteed to execute a task (possibly with high probability) irrespective of the realization of the joint space error is called a robust IK solution. In this paper we formulate and solve the robust inverse kinematics problem for redundant manipulators with actuation uncertainties (errors). We also present simulation and experimental results on a $7$-DoF redundant manipulator for two applications, namely, a pre-grasp positioning and a pre-insertion positioning scenario. Our results show that the robust IK solutions result in higher success rates and also allows the robot to self-evaluate how successful it might be in any application scenario.}
\end{abstract}


\section{INTRODUCTION}
\label{sec: intro}
\noindent
A fundamental problem in many robotics tasks is to move the end effector of a manipulator to a desired configuration (position and orientation). For example, in grasping, the end effector is usually moved to a pre-grasp configuration before closing the fingers. In assembly operations like peg-in-a-hole operations, the end effector holding a part is usually moved to a pre-insertion configuration from where the insertion of the peg in the hole is attempted. One key factor that dictates the success of operations like grasping or the peg-in-a-hole insertion is the accuracy of the placement of the end effector in the pre-grasp or pre-insertion configuration. Because of the inherent inaccuracy in actuation, it is usually not possible to place the end effector at the desired configuration exactly. Thus, success of operation depends on the amount of inaccuracy that can be tolerated for the given task and whether robot can place its end effector within that error margin. 
\noindent
Furthermore, given a particular accuracy requirement for a given task, a question of interest is for the robot to evaluate its capability on whether it can place its end effector so that it is highly likely that the robot can perform the subsequent operations successfully. The {\em goal of this paper is to develop algorithms for the robot to select its joint angles such that in spite of actuation and/or sensing errors, the robot can determine if it can place its end effector within the prescribed error margin and also compute the solution that is robust to the actuation errors}.


\begin{figure}[!t]
	\centering
	\includegraphics[scale=.33]{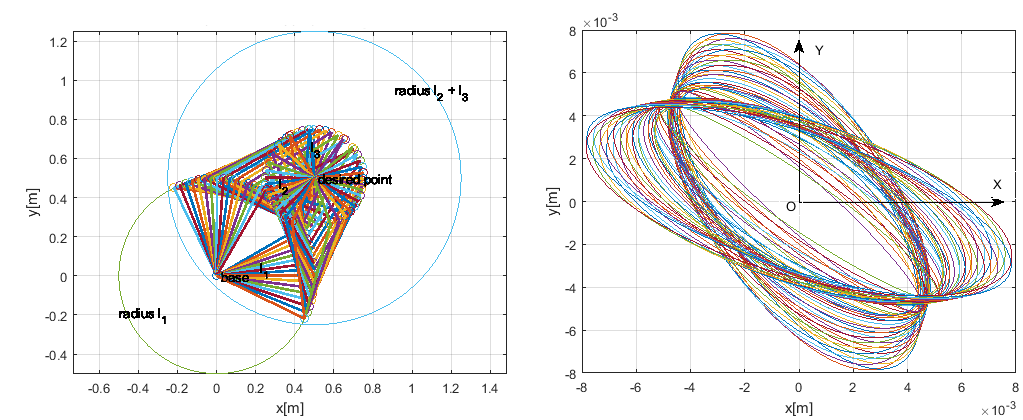}
	\caption{(Left) multiple ik solutions of $3$R robot for a reaching task. (Right) task space error sets corresponding to IK solutions in left figure w.r.t. End-effector frame, obtained by propagating joint space error set. All error sets are centered around $\bm{O}$ (desired position) but with different size and orientations because same error set in joint space mapped to different error sets in task space for different IK solution}
	\label{fig: 3r_all}     
\end{figure}

Computing the joint angles given end effector (tool) configuration is known as inverse kinematics (IK) problem which can be formally defined as follows: {\em Given a desired position and orientation of the end effector of the manipulator, compute the corresponding joint angles.} It is well known that, in general, the IK problem has multiple solutions, and for redundant robots, the IK problem has infinitely many solutions. All the solutions are equivalent in a sense that end effector goes to the same configuration if each joint can be rotated exactly as desired. However, if there is execution error, the effect of the joint error in the task space is not identical. For example, consider the $3R$ manipulator in Figure\ref{fig: 3r_all}, where the goal is to move the end effector to a given position in the $xy$-plane. Since there are only two variables in the task space (namely, $(x, y)$ coordinates of the end effector), and the $3R$ robot has $3$ degrees-of-freedom (DoF), the robot is redundant with respect to the positioning task. Hence number of IK solutions is infinity and some of them are shown in Fig\ref{fig: 3r_all}. 

\begin{figure}[h]
	\centering
	\includegraphics[scale=.33]{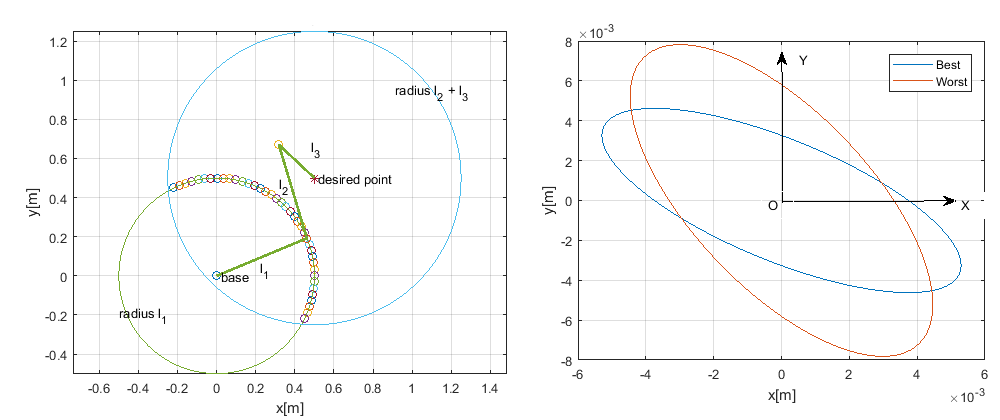}
	\caption{The task space error sets corresponding to Robust (blue) and worst (red) joint configurations for a task that requires minimized error along $Y$ axis. Blue ellipse has smaller projection length as compared to the projection of red ellipse on to $Y$ axis}
	\label{fig: 3r_best_case}
\end{figure}

Now let us assume that that the joint space error is bounded within  a $3$ dimensional ball of radius $r = 0.005$. By propagation of error to task space, we obtain an error ellipse (see Figure\ref{fig: 3r_all}) for each joint space configuration. 
As is evident from the Fig\ref{fig: 3r_all}, the task space error sets for different IK solutions are different. Therefore, if the positioning task demands minimum error in $x$-direction, then IK solution corresponding to the red error-ellipse is better than the IK solution corresponding to blue error-ellipse (see Figure\ref{fig: 3r_best_case}). However if criterion for the positioning task of the end effector is to have minimum error along $y$-direction, then IK solution corresponding to the blue error-ellipsoid is better. Furthermore, if we can tolerate an error of $5$ mm (say) along the $y$-axis, the IK solution corresponding to blue ellipse will always guarantee it, irrespective of the joint space error (as long as it is bounded within the ball of radius $0.005$). Thus, the blue solution is {\em robust} to the task-specific requirement whereas the red solution is not.

\noindent

The above planar example illustrates the need for developing methods to analyze the propagation of the joint space errors and find IK solutions that are {\em robust} to the actuation errors.
The robust IK problem is defined as follows: {\em Given a desired tool configuration, a joint space error bound, and a tolerance parameter $\epsilon$, compute a joint space configuration such that the error between the actual tool configuration and the desired configuration is less than $\epsilon$ (with high probability)}. 
Although IK is a classical problem in robotics and there has been recent work on
propagation of joint space errors to task space of robotic manipulators  in~\cite{chirikjian2001,Barfoot2014}, to the best of our knowledge, there is no work on finding IK solutions that are robust to actuation errors. Algorithms to compute robust IK solutions can enable a robot to determine whether it can successfully accomplish a task based on given error tolerance. For example, for the planar $3R$ robot with the joint space error bound specified above, if the requirement was to have a guaranteed error along the $y$-axis of at most $2$ mm, none of the IK solutions could have produced it. Thus, when there is no solution to the robust IK problem and the robot can know that it cannot do the task reliably.

Although the robust IK problem is a feasibility problem, we will formulate and solve the minimization version of the problem, which is more general. Here, we want to compute the joint space solution that minimizes the maximum error from the desired configuration, (irrespective of the realization of the joint space error). We will call this solution the IK solution with best task-specific error characteristics or simply the {\em best IK solution}. Assuming that the joint space errors are small, we present an algorithm to compute the {\em best} IK solution as well as robust IK solution (if one exists). {\em The formulation of the robust (or best) IK problem and it's solution algorithm is the key contribution of the paper}.

%
%
We use a geometric model (e.g., $n$ dimensional ball) to bound joint space uncertainty, which can be used to have both set-theoretic and probabilistic interpretation of uncertainty models. Since there is no bi-invariant metric on the space of rigid body configurations, $SE(3)$, we use a task-specific metric as a measure of error between two configurations. This involves selecting a task-specific weighting of rotational error and translational error. We show that, using task-specific error measure, the optimization problem for computing best IK solution can be decoupled into two separate problems of optimizing over the space of rigid body translations ($\mathbb{R}^3$) and space of rigid body rotations ($SO(3)$). We demonstrate the usefulness of computing the robust IK solution both in simulation and experiment using the Baxter robot from Rethink Robotics. We use two example scenarios, one on moving to a pre-grasp configuration for grasping with a parallel jaw gripper and the other on moving to a pre-insertion configuration bi-manual peg-in-hole assembly. 

\section{RELATED WORK}
\label{sec: rel_wrk}
\noindent
The error in positioning the end effector of a robot arm arises due to uncertainties in actuation, sensing, robot model, and control algorithms. Irrespective of the source of uncertainty, we can broadly categorize uncertainty into two basic classes, namely {\em static errors} and {\em random errors}. Errors in link-lengths, offset lengths, and/or origin of the joints that are not known precisely introduce constant biases in end effector configuration which are often denoted as \textit{static errors}. They do not change over time and hence can be estimated offline and compensated during calibration process of the robot. Works on estimating these parameters using end effector pose error between kinematics model and actual pose can be found in~\cite{Meggiolaro, Mavroidis1997, Wu1983, Wu1984, Veitschegger1986, Chen1984, Chen1987, mooring1991fundamentals, omodei2001calibration,nof1999handbook}. To accurately collect end-effector pose data to estimate error parameters,~\cite{Yin2012} used an external laser tracker. Static pose error minimization for calibration using partial pose information can be found in~\cite{Goswami1993}. Representation of spatial uncertainty in the context of robotics can be found in the classic work by~\cite{Smith1986, Su1992} and recently in~\cite{Barfoot2014}. Authors in~\cite{Barfoot2011} presented a method to map rotation errors from roll-pitch-yaw to quaternion representation of rotations.   

The second kind of error corresponds to random actuation and sensing errors realized in execution time of the robot. They implicitly affect accuracy in joint rotations which in turn affects accuracy at the end-effector of a manipulator. This second kind of error source in positioning tasks is the motivation behind our work. A group theoretic approach to propagate joint space random actuation error into end-effector space has been presented in~\cite{Wang2006}. In~\cite{Wang2006}, the authors present a method to obtain error covariance at the end of each individual link in closed form due to errors in desired joint configurations. By repeating this procedure sequentially for each link of a manipulator they obtain final error covariance at the end effector. To capture the effect of large joint errors on error covariance authors in~\cite{Wang2008} presented a second order theory of error propagation. In essence their method relies on evaluating desired and erroneous poses at each joints for a few discrete samples of joint errors so that individual covariances at each joint can be calculated and finally combine them in sequential manner using proposed closed form covariance propagation formula to obtain final error covariance at the last distal frame. 

None of the methods for tackling the biases or the random errors, address the problem of computing the inverse kinematics configurations for which the propagated error is sufficient for the task at hand. Our goal in this paper is to tackle this IK problem. Thus, unlike~\cite{Wang2006, Wang2008} our method does not rely on individual error samples and corresponding frame by frame error covariance computation. We assume small joint errors along with linearized model of forward position and rotation kinematics to propagate a geometric description of the joint space error set into task space error set. Obviously~\cite{Wang2006, Wang2008} are more effective if joint space errors are large and no prior knowledge about joint error bounds is available. 

\section{MATHEMATICAL PRELIMINARIES}
\label{sec:math_prelim}
\noindent
In this section, we present notations and definitions that will be used throughout the paper. Let ${\mathbb  R}^n$ be the real Euclidean space of dimension $n$,  ${\mathbb  R}^{m \times n}$ be the set of all $m \times n$ matrices with real entries. The set of all joint angles, ${\mathcal J}$, is called the {\em joint space} or the configuration space of the robot. In this paper ${\mathcal J} \subset {\mathbb  R}^7$, since we are using a $7$ Degree-of-Freedom ($7$-DoF) manipulator with joint limits. 

Let $SO(3)$ be the Special Orthogonal group of dimension $3$, which is the space of all rigid body rotations. Let $SE(3)$ be the Special Euclidean group of dimension $3$, which is the space of rigid motions (i.e., rotations and translations). $SO(3)$ and $SE(3)$ are defined as follows~\cite{MLS1994}: 
$SO(3) = \{{\bf R} \subset {\mathbb R}^{3 \times 3}| {\bf R}^T{\bf R} = {\bf R}{\bf R}^T = {\bf I}, |{\bf R}| = 1 \}$,
$ SE(3) =  SO(3) \times {\mathbb R}^3 = \{({\bf R},{\bf p}) | {\bf R} \in SO(3), {\bf p} \in {\mathbb R}^3 \} $
where $|{\bf R}|$ is the determinant of ${\bf R}$ and ${\bf I}$ is a $3 \times 3$ identity matrix.  The set of all end effector configurations is called the {\em end effector space} or {\em task space} of the robot and is a subset of $SE(3)$.


\noindent
{\bf Unit Quaternion Representation of $SO(3)$}:
Unit quaternions are a singularity free representation of $SO(3)$.
A quaternion is a tuple $\bm{q} = (\eta, \epsilon_x, \epsilon_y, \epsilon_z$) which includes a vector $\bm{\epsilon} \in \mathbb{R}^3$ with components $\epsilon_x$, $\epsilon_y $, $\epsilon_z$ and a scalar $\eta$.
For a unit quaternion $\|{\bf q}\| = 1$. In our paper we extensively make use of vector representation of unit quaternions, $\bm{q} = [\eta \quad \bm{\epsilon}^T]^T$ with its conjugate $\bm{q}^{-1} = [\eta \quad -\bm{\epsilon}^T]^T$. Rotation about an axis $\bm{\omega}$ with angle $\phi$ is a unit quaternion represented as, $\bm{q}(\bm{\omega}, \phi) = [\cos \phi/2 \quad \bm{\omega}\sin \phi/2]$. Let $\bm{\epsilon}^\times \in \mathbb{R}^{3 \times 3}$ denote \textit{skew symmetric matrix} of vector $\bm{\epsilon} \in \mathbb{R}^3$.

\section{PROBLEM FORMULATION}
\label{sec: problem_formulation}
\noindent
In this section we formally describe the problem of computing robust-IK and pose the problem as a constrained optimization problem. We do it in three steps. First, we describe our set-theoretic approach to model joint space uncertainty. Then, we show the steps to propagate uncertainty from joint-space to task-space through a linearized position and rotation kinematics map. After second step we will be able to find task error sets for position and orientation independently. In third step we formulate the task dependent robust-IK problem. Before we move forward, we define the term \textit{error bound} as following: \textit{by error-bound we mean maximum possible error in position or orientation one may encounter most of the time (with very high probability) due to actuation or sensing error in joint space}.

\subsection*{Uncertainty Modeling in Joint Space: }
\noindent
We use a geometric model of uncertainty in the joint space. More specifically, we model the joint space error as a ball in $\mathbb{R}^n$ centered at the desired joint angles. The ball can be interpreted as either a set-theoretic model of uncertainty or a probabilistic model of uncertainty. Depending on the interpretation of the uncertainty model, the results can be given a worst case interpretation or a probabilistic interpretation.

In the set-theoretic model, the ball represents the $2$-norm of the joint errors, and the assumption is that the joint errors always lie within this set and we have no other knowledge that can be used to define a probability measure on this set. 

In the probabilistic model, the ball represents the probability mass that the uncertain configuration is within the ball assuming a Gaussian probability measure. The radius of the ball corresponds to the confidence level with which we want to know whether the uncertain configuration will lie within the ball. Thus, the actuation uncertainty, $\bm{\delta \Theta}$, is a multivariate Gaussian with zero mean and known $n \times n$ covariance matrix $\left( \bm{\Sigma}\right) $, i.e., $\bm{\delta \Theta} \sim \mathcal{N} \left( {\bf 0}, \bm{\Sigma} \right)$. The ball model implies that the covariance matrix is diagonal or the error in the joints are uncorrelated and the variance of all the joints are identical, i.e., $\bm{\Sigma} = \sigma^2\mathbb{I}$, where $\sigma$ is the standard deviation and $\mathbb{I}$ is the $n\times n$ identity matrix.
Thus, the uncertainty set in the joint space is
\begin{equation}
\label{eq:js_error_ball}
\bm{\delta \Theta}^T \bm{\delta\Theta} \leq c \quad \text{where} \quad c = (k \sigma)^2 
\end{equation}  
considering error up to $k$ standard deviations about zero mean.
\begin{remark}
Please note that the ball model is for ease of exposition only. We could have used a more general ellipsoidal model of the uncertainty, with the covariance matrix non-diagonal. The problem formulation and the solution techniques applies in this case also, although the expressions for the propagated uncertainty set becomes more complicated.
\end{remark}

We assume that the joint actuation errors are sufficiently small such that a linear approximation of the forward kinematics function about the desired position and orientation can be used to propagate joint space error. To propagate joint space error (as in equation~\eqref{eq:js_error_ball}) into task space, we need to have the mapping from $\bm{\delta \Theta} \in \mathbb{R}^n$ to $\bm{\delta T} \subset SE(3)$ where $\bm{\delta T} = (\bm{\delta X}, \bm{\delta q})$, $\bm{\delta X} \in \mathbb{R}^3$, $\bm{\delta q} \in SO(3)$. To find expression of $\bm{\delta q}$ in terms of $\bm{\delta \Theta}$, we use first order Taylor's series approximation of \textit{forward rotation kinematics function} $\bm{q}(\bm{\Theta}) : \mathbb{R}^N \rightarrow SO(3)$. Similarly we derive the expression for position error term i.e., $\bm{\delta X} \in \mathbb{R}^3$ by linearizing {\em forward position kinematics function} $\bm{F}(\bm{\Theta}):\mathbb{R}^N\rightarrow\mathbb{R}^3$.  
Since there is no bi-invariant metric available~\cite{belta2002} to quantify error in $SE(3)$, we optimize two metrics defined over $\mathbb{R}^3$ and $SO(3)$ independently. To compute \textit{robust IK} we optimize an objective obtained by weighted sum of error metrics defined over $\mathbb{R}^3$ and $SO(3)$.

\subsection{Error Propagation from joint space to $\mathbb{R}^3$}
\noindent
Suppose $\bar{\bm{\Theta}} \in \mathbb{R}^n$ denotes the nominal joint angle vector of a $n$ DoF manipulator that takes the end-effector to desired position $\in \mathbb{R}^3$. Due to joint angle errors $\bm{\delta \Theta}\in \mathbb{R}^n$, actual position of end effector will be $\bm{F}\left(\bar{\bm{\Theta}} + \bm{\delta \Theta}\right)$ instead of desired position $\bm{F}\left(\bar{\bm{\Theta}}\right)$. The Taylor's series approximation of $\bm{F}(\bar{\bm{\Theta}}+\bm{\delta \Theta})$ gives,
\begin{equation}
\label{eq: Taylors}
\bm{F}\left(\bar{\bm{\Theta}} + \bm{\delta \Theta}\right) = \bm{F}
\left(\bar{\bm{\Theta}}\right) + \frac{\partial \bm{F}}{\partial \bm{\Theta}}|_{\bar{\bm{\Theta}}} \bm{\delta \Theta} + \bm{\mathcal{O}}\left(\bm{\delta \Theta}^2\right)
\end{equation}
The partial derivative  $\frac{\partial \bm{F}}{\partial\bm{\Theta}}|_{\bar{\bm{\Theta}}} \in \mathbb{R}^{3 \times n}$ corresponds to the first three rows of {\em manipulator Jacobian}~\cite{MLS1994, Craig}. Denoting $\frac{\partial \bm{F}}{\partial\bm{\Theta}}|_{\bar{\bm{\Theta}}}$ by $\bm{J}_p\left(\bar{\bm{\Theta}}\right) \in \mathbb{R}^{3 \times n}$ (\textit{position Jacobian}), \textit{position error vector} is
\begin{equation}
\label{eq:error_Taylors}
\bm{\delta X} = \bm{F}(\bar{\bm{\Theta}} + \bm{\delta \Theta}) - \bm{F}(\bar{\Theta}) \approx  \bm{J}_p \bm{\delta \Theta}
\end{equation}
where $\bm{J}_p\in\mathbb{R}^{3 \times n}$, $n > 6$. Therefore,
\begin{equation}
\label{eq:vel_inv1}
  \bm{\delta \Theta} = \bm{J}_p^T\left[\bm{J}_p\bm{J}_p^T\right]^{-1} \bm{\delta X} 
\end{equation}
Substituting~\eqref{eq:vel_inv1} into~\eqref{eq:js_error_ball} and simplifying, we obtain
\begin{equation}
\label{eq:err4}
\bm{\delta X}^T \left[\bm{J}_p\bm{J}_p^T\right]^{-1}\bm{\delta X} \leq c
\end{equation}
Equation~\eqref{eq:err4} describes the set of position errors of the end effector due to the error in the joint space. Geometrically, the set of position errors is an ellipsoid.

\subsection{Error Propagation from Joint Space to $SO(3)$}
\noindent
As before, let $\bar{\bm{\Theta}} \in \mathbb{R}^n$ denote nominal joint angle vector that takes end-effector to desired orientation $\bm{q}\left( \bar{\bm{\Theta}} \right)$. Because of random joint error $\bm{\delta \Theta}$, actual orientation of end effector becomes $\bm{q}\left( \bar{\bm{\Theta}} + \bm{\delta \Theta} \right)$. Using first order Taylor's series approximation we get error quaternion $\bm{\delta q}$ as,
\begin{equation}
\label{eq: barfoot_firstord}
	\bm{\delta q} = \bm{q}(\bar{\bm{\Theta}} +  \bm{\delta\Theta}) - \bm{q}(\bar{\bm{\Theta}}) \approx\frac{\partial \bm{q}(\bm{\Theta})}{\partial 			\bm{\Theta}} \mid_{\bar{\bm{\Theta}}}\bm{\delta\Theta} + \left( \bm{\delta\Theta} \right)^2
\end{equation}
\noindent
Following~\cite{Barfoot2011}, for serial chain manipulators, we can show
\begin{eqnarray}
\label{eq: barfoot_firstord_compact}
\bm{\delta q} & = & \frac{\partial \bm{q}_r(\bm{\Theta})}{\partial \bm{\Theta}}\bm{\delta \Theta} = \frac{1}{2}\bm{H}^T \bm{J}_r\bm{\delta \Theta}\\
\text{where }
\bm{H}(\bm{q}) & = &
\begin{bmatrix}
-\bm{\epsilon} & \eta\mathbb{I} - \bm{\epsilon}^\times 
\end{bmatrix} \\ \nonumber
\bm{J}_r & = &
\begin{bmatrix}
\bm{\omega}_1 & \bm{R}_1\bm{\omega}_2  &  \bm{R}_1\bm{R}_2\bm{\omega}_3 & \cdots & \prod_{i=1}^{n-1}\bm{R}_i\bm{\omega}_n
\end{bmatrix}
\end{eqnarray}
where $\bm{R}_i$ is the rotation matrix of $i^{th}$ frame and $\bm{\omega}_i$ is the axis of rotation of the same. Utilizing the fact that $\bm{H}\bm{H}^T=\mathbb{I}_{3 \times 3}$ and Eq~\eqref{eq: barfoot_firstord_compact} we obtain the relationship between $\bm{\delta q}$ and $\bm{\delta \Theta}$ as,
\begin{equation}
    \label{eq:dth_dq}
    \bm{\delta \Theta} = 2\bm{J}_r^T(\bm{J}_r\bm{J}_r^T)^{-1}\bm{H}_d\bm{\delta q}
\end{equation}
where $\bm{H}_d = \bm{H}(\bm{q}_d)$.
Substituting $\bm{\delta \Theta}$ from Eq~\eqref{eq:dth_dq} into Eq~\eqref{eq:js_error_ball} we get \textbf{uncertainty set} of orientation in task space as follows,
\begin{equation} 
    \label{eq:uncertainty_SO3}
    \bm{\delta q}^T \bm{H}_d^T (\bm{J}_r \bm{J}_r^T)^{-1}\bm{H}_d \bm{\delta q} \leq c/4
\end{equation}
\noindent
Thus we have propagated error set from joint to rotation task space as in equation~\eqref{eq:uncertainty_SO3}. Next we formulate the optimization problem to compute robust-IK solution for a given task defined in end-effector space.

\subsection{Task-dependent Robust IK Solution}
\noindent
Let $\bm{g}_d \in SE(3)$ be the desired tool frame configuration, where $\bm{g}_d = (\bm{x}_d, \bm{q}_d)$, with $\bm{x}_d \in \mathbb{R}^3$ denoting the desired position and $\bm{q}_d$ denoting the unit quaternion representation of the desired orientation of the tool frame. Let $\mathbb{M}$ be a task-specific metric or measure of distance between two configurations in $SE(3)$. Note that technically, there is no bi-invariant metric in $SE(3)$~\cite{belta2002}, and the definition of a metric involves a choice of a weight relating the rotational error in $SO(3)$ to the translational error in $\mathbb{R}^3$. Thus, we assume $\mathbb{M} = \mathbb{P} + \lambda \mathbb{O}$, where $\mathbb{P}$ is a metric defined on $\mathbb{R}^3$ and $\mathbb{O}$ is a metric defined on $SO(3)$. The weight $\lambda$ is a task-dependent parameter. Some examples of the metric $\mathbb{M}$ of interest are given in the later sections. Note that due to error in execution, any IK solution will result in an end effector configuration that is different from $\bm{g}_d$. 

Let $\mathcal{N}(\bm{g}_d)$ be a neighborhood of the desired configuration $\bm{g}_d$, such that any end effector configuration $\bm{g} \in \mathcal{N}(\bm{g}_d)$ is an acceptable solution for the task at hand. Let $\epsilon$ be a task-dependent tolerance. Then $\mathcal{N}(\bm{g}_d) = \{\bm{g}: \mathbb{M} (\bm{g}, \bm{g}_d) \leq \epsilon\}$. Formally a robust IK solution is defined as {\em a joint angle vector for which the end effector lies within the neighborhood $\mathcal{N}(\bm{g}_d)$ with high (user prescribed) probability irrespective of the realization of the random joint space error during task execution}. Note that for a given $\bm{g}_d$ there may be multiple IK solutions that are robust. It may also be possible that there are no robust solutions. Therefore we pose the robust IK problem as an optimization problem (instead of a feasibility problem) as follows:
\begin{eqnarray}\nonumber
	\label{eq: main_opt_problem}
	\underset{\bm{\Theta} \in \mathbb{R}^n}{\text{argmin\quad}}\underset{\bm{x}, \bm{q}}{\text{max\quad}} &  & \mathbb{M}\\
	\text{subject to           } & & \bm{g}_{st}(\bm{\Theta}) = \bm{g}_{d} \\ \nonumber
	& &\bm{\delta x}^T \left[\bm{J}_p\bm{J}_p^T\right]^{-1}\bm{\delta x} \leq c \\ \nonumber
	& & \bm{q}^T\bm{q} = 1 \\ \nonumber
	& &\bm{\delta q}^T\left[\bm{H}_d^{T}\left(\bm{J}_r\bm{J}_r^T\right)^{-1}\bm{H}_d\right]\bm{\delta q}  \leq  \frac{c}{4}
\end{eqnarray}
where $\bm{x} = \bm{x}_d + \delta\bm{x}$, $\bm{q} = \bm{q}_d + \delta\bm{q}$. If $\mathbb{M} \leq \epsilon$, then the optimal solution is a robust IK, which we will denote by $\bm{\Theta}^*$. Otherwise, there is no robust IK solution. In the above formulation, all the four constraints are dependent on joint vector $\bm{\Theta}$ but for brevity we have not showed them explicitly. The first constraint ensures that $\bm{\Theta}^*$ would indeed be a joint solution of the manipulator that would take the end effector to desired configuration $\bm{g}_d \subset SE(3)$. The second constraint ensures that position error would lie inside its respective error set as described by equation~\eqref{eq:err4}. The fourth constraint ensures rotation error will lie inside the error set as in equation~\eqref{eq:uncertainty_SO3}. 
The third constraint makes sure that computed $\bm{q}$ is a unit quaternion. The parameter $c$ is from equation~\eqref{eq:js_error_ball} and depends on the confidence level considered while bounding joint space error.

\section{SOLUTION APPROACH}
\label{sec:solution_approach}
\noindent
In Eq~\eqref{eq: main_opt_problem}, since $\mathbb{M} = \mathbb{P} + \lambda \mathbb{O}$, the  objective is separable. The constraints are either in $\bm{x}$ or in $\bm{q}$. Therefore the optimization problem in Eq~\eqref{eq: main_opt_problem} can be written as two decoupled optimization problems as follows:
\begin{eqnarray}
\label{eq: opti_position_only}
\underset{\bar{\bm{\Theta}}}{\text{argmin\quad}}\underset{\bm{\delta x}}{\text{max\quad}} &  & \mathbb{P}\\ \nonumber
\text{subject to           }
& &\bm{\delta x}^T \left[\bm{J}_p(\bar{\bm{\Theta}})\bm{J}_p^T(\bar{\bm{\Theta}})\right]^{-1}\bm{\delta x} \leq c \\ 
\label{eq: opti_orientation_only}
\underset{\bar{\bm{\Theta}}}{\text{argmin\quad}}\underset{\bm{q}}{\text{max\quad}} &  & \mathbb{O}\\ \nonumber
\text{subject to           }
& & \bm{q}^T \bm{q} = 1 \\ \nonumber
& &
\bm{\delta q}^T \left[\bm{H}_d^{T}\left(\bm{J}_r(\bar{\bm{\Theta}})\bm{J}_r^{T}\right)^{-1}(\bar{\bm{\Theta}})\bm{H}_d\right]\bm{\delta q}  \leq  \frac{c}{4}
\end{eqnarray}

\noindent
Now we discuss methods of solving the inner maximization problem in Eq~\eqref{eq: opti_position_only} and~\eqref{eq: opti_orientation_only} efficiently for one particular $\bar{\bm{\Theta}}$. Output of these inner max problems are error bounds in position or orientation respectively. After solving this inner max problem for multiple IKs, the outer min problem is just finding the minimum of the computed error bounds. 

\subsection{Computing position error bound for a given $\bar{\bm{\Theta}}$}
\noindent
The problem of finding error bounds for a particular $\bar{\bm{\Theta}}$ that will satisfy position error ellipsoid constraint in task space is defined in Eq~\eqref{eq: main_position_opti}.
\begin{eqnarray}
\label{eq: main_position_opti}
\underset{\bm{\delta x}}{\text{maximize\quad}} &  & \mathbb{P}\\ \nonumber
\text{subject to           } 
& &\bm{\delta x}^T \left[\bm{J}_p(\bar{\bm{\Theta}})\bm{J}_p^{T}(\bar{\bm{\Theta}})\right]^{-1}\bm{\delta x} \leq c
\end{eqnarray}
Please recall that origin of the error ellipsoid is desired tool position (see figures~\ref{fig: 3r_all} or~\ref{fig: 3r_best_case}) whereas any other point lying inside the error set is a deviation from the desired position. The choice of metric $\mathbb{P}$ may vary depending on whether the task demands to minimize position error in $\mathbb{R}^3$ or any of its sub-spaces. Next we discuss on computing position error bounds in $\mathbb{R}^3$ and its lower dimensional sub-spaces, {i.e.}, $\mathbb{R}$ and $\mathbb{R}^2$ respectively.\\
\noindent
\textbf{Computing position error bound in $\mathbb{R}^3$: }
Maximum possible position error or position error bound in this case would be the distance from the center of the task space position error ellipsoid (which is desired tool position) to the furthest point in the error ellipsoid. Following the fact the maximum eigenvalue of an ellipsoid describes the distance of the furthest point from its center, the optimization problem in equation~\eqref{eq: opti_position_only} is reduced to as in equation~\eqref{eq: pst_opt_eigen_max}.
\begin{equation}
\label{eq: pst_opt_eigen_max}
\mathbb{P}^* = \underset{\lambda}{\text{max\quad}} \text{eig} \left(c\left[\bm{J}_p(\bar{\bm{\Theta}})\bm{J}_p(\bar{\bm{\Theta}})^T\right]\right)
\end{equation}
where $\mathbb{P}^*$ denotes $max(\mathbb{P})$ and {\em eig} a function that computes all eigenvalues of an input matrix. $\bm{J}_p$ and $\bar{\bm{\Theta}}$ are defined as before.\\
\noindent
\textbf{Computing position error bound in along a direction: }
Computing position error bound along a direction becomes necessary for tasks such as, drawing a horizontal line by a robot arm. In that case the arm needs minimum deviation of end-effector along vertical direction. Therefore position error bound for this case would be projection length of error ellipsoid onto the vertical axis. In general if we want to minimize error along a direction $\bm{v} \in \mathbb{R}^3$ then position error bound would be $\mathbb{P}^* = \left| \frac{\bm{L}^{-1}_c\bm{v}}{\bm{v}^T\bm{v}} \right|$, where $\bm{L}_c$ is lower-triangular matrix obtained by Cholesky decomposition of $\frac{1}{c}\left[\bm{J}_p\bm{J}_p^T\right]^{-1}$.

\noindent
\textbf{Computing position error bound in $\mathbb{R}^2$ or along a plane: }
\label{sec: opti_dir}
In this case position error bound is the projected area of position error ellipsoid onto the plane along which error minimization is required. Suppose the given plane equation is defined as $PL \equiv \{\bm{x}|\bm{x}=\bm{Tt}\}$ where $\bm{T}$ is the basis matrix defining the plane, then position error bound $\mathbb{P}^* = det\left| \left( \bm{TT}^T \right)^{-T} \bm{L}_c \bm{L}_c^{T} \left( \bm{TT}^T \right)^{-1}\right|$ where $\bm{L}_c$ is defined as before.


\subsection{Computing rotation error bound for given $\bar{\bm{\Theta}}$}
\begin{figure}
    \centering
    \includegraphics[scale=0.2]{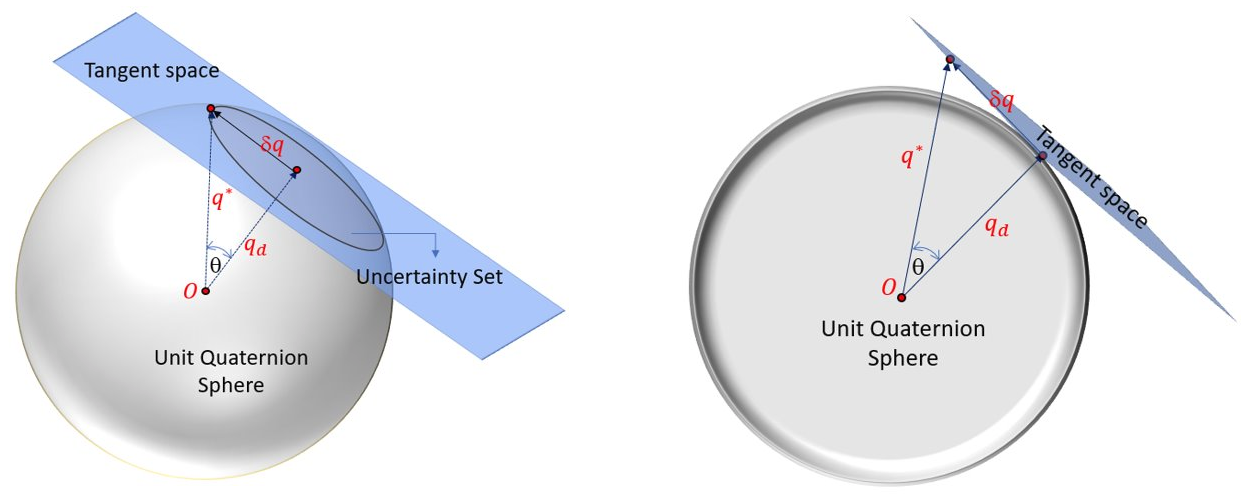}
    \caption{From left: Unit-quaternion sphere with uncertainty set(ellipsoid) in its tangent space. A vertical slice of uni-quaternion sphere in the plane defined by $\bm{q}_d$ and $\bm{\delta q}$}
    \label{fig:geometric_visualization}
\end{figure}

\noindent
To compute rotation error bound for an IK, say $\bar{\bm{\Theta}}$, we need to solve the inner max-problem as in equation~\eqref{eq: opti_orientation_only}. If we consider orientation error metric to be $\mathbb{O}=\bm{q}^T\bm{q}_d$ following~\cite{Huynh2009}, then the inner max problem turns into a min problem since $\mathbb{O}$ in this case would define \textit{cosine} of the included angle. Then inner optimal problem now turns out as in equation~\eqref{eq: rotation_opt_problem0}.
\begin{eqnarray} \nonumber
\label{eq: rotation_opt_problem0}
\underset{\bm{q}}{\text{min}} & & \bm{q}_d^T\bm{q}\\
\text{s.t.           }
& & \bm{q}^T\bm{q}  = 1 \\ \nonumber
& & \bm{\delta q}^T\bm{H}_d^{T} \left(\bm{J}_{r}\bm{J}_{r}^{T}\right)^{-1}\bm{H}_d \bm{\delta q}  \leq  \frac{c}{4}
\end{eqnarray}
The optimization problem in equation~\eqref{eq: rotation_opt_problem0} is non-convex because of the strict quadratic equality unit quaternion constraint. Nonconvex problems are hard in general and optimization solvers ends up to local minimum instead of global minimum while solving them. We can relax the problem by ignoring unit quaternion constraint. However to ensure that we still minimize the included angle between $\bm{q}_d$ and $\bm{q}$, we need to scale the objective in Eq~\eqref{eq: rotation_opt_problem0} with the length of $\bm{q}$. The relaxed optimization problem is presented in Eq~\eqref{eq:main_rotation_optimization}.

\begin{eqnarray} 
    \label{eq:main_rotation_optimization}
    \bm{q}^* &=& \underset{\bm{q}}{\text{min }} \bm{q}^T\bm{q}_d/||\bm{q}||\\ \nonumber
    & &\text{s.t. }\bm{\delta q}^T \bm{H}_d^T (\bm{J}_r \bm{J}_r^T)^{-1}\bm{H}_d \bm{\delta q} \leq c/4
\end{eqnarray}
We have presented a schematic in figure~\ref{fig:geometric_visualization} to visualize optimization problem in  equation~\eqref{eq:main_rotation_optimization}. The $3D$ ball represents a unit quaternion sphere (in gray) whereas blue plane represents the tangent space of unit quaternion sphere touching it at $\bm{q}_d$. The ellipse, lying on the tangent space is the uncertainty set centered around $\bm{q}_d$. Any vector lying inside this ellipsoid with the starting point fixed at $\bm{q}_d$ will represent random rotation error $\bm{\delta q}$ with respect to $\bm{q}_d$.
Assuming $\bm{H}_d\bm{\delta q} = \bm{v}$ we get the expression of uncertainty set as $\bm{v}^T(\bm{J}_r \bm{J}_r^T)^{-1} \bm{v} \leq c/4$. Further we can compute $\bm{q}$ from definition of $\bm{v}$ as following,
\begin{equation}
    \label{eq:q_in_terms_v}
    \bm{H}_d\bm{\delta q} = \bm{v} \implies \bm{\delta q} = \bm{H}_d^T \bm{v} \implies \bm{q} = \bm{q}_d + \bm{H}_d^T \bm{v}
\end{equation}
\noindent
Utilizing Eq~\eqref{eq:q_in_terms_v} and the fact that $\bm{H}_d\bm{q}_d = 0$ we can write optimization objective in Eq~\eqref{eq:main_rotation_optimization} as, $\bm{q}^T\bm{q}_d/||\bm{q}|| = 1/\sqrt{1+\bm{v}^T\bm{v}}$. Then optimization problem in Eq~\eqref{eq:main_rotation_optimization} can be written as,
\begin{eqnarray} \nonumber
    \label{eq:rotation_optimization_v_2}
    \bm{v}^* &=& \underset{\bm{v}}{\text{max }} \bm{v}^T\bm{v}\\
    & &\text{s.t. } \bm{v}^T (\bm{J}_r \bm{J}_r^T)^{-1}\bm{v} \leq c/4
\end{eqnarray}


\noindent
The optimization problem in equation~\eqref{eq:rotation_optimization_v_2} represents eigenvalue problem. Mathematically $\bm{v}^* = \frac{1}{2}\sqrt{c \lambda_{max}}\bm{V}_{max}$
where $\lambda_{max}$ is maximum eigenvalue of $[\bm{J}_r\bm{J}_r^T]$ and $\bm{V}_{max}$ is the eigenvector associated to $\lambda_{max}$. Once we have computed $\bm{v}^*$ we can compute bounding quaternion $\bm{q}^*$ by following equation~\eqref{eq:q_in_terms_v} as,
\begin{equation}
    \label{eq:q_star}
    \bm{q}^* = \bm{q}_d + \bm{H}_d^T \bm{v}^*/||\bm{q}_d + \bm{H}_d^T \bm{v}^*||
\end{equation}
Once we obtain $\bm{q}^*$, we find rotation error bound as,
\begin{equation}
\label{eq:rotation_error_measure}
    \mathbb{O} = \arccos \bm{q}_d^T\bm{q}^*
\end{equation}
\noindent
In Algorithm~\ref{alg:SMB} we have presented the steps required to compute robustIK in concise manner for the ease of implementation.

\section{ALGORITHM}
\noindent
Algorithm~\ref{alg:SMB} takes desired end-effector configuration $\bm{g}_d$, $c=(k\sigma)^2$, metric weighing factor $\lambda$ and allowable error tolerance $\epsilon$ as input. Output of the algorithm is robust IK, $\bm{\Theta}^*$. Line 1 separates desired position $\bm{p}_d$ and rotation $\bm{q}_d$ from $\bm{g}_d$. Line 2 computes $M$ IK solutions~\cite{Sinha2019} and stores them in $\bm{\Gamma}$. Then we iterate over IK solution set between lines 3 and 8. For each IK solution we compute error bounds for position (line $5$), rotation (line $6$) and finally weighted error bound in line $7$. If computed error bound is lesser or equal to $\epsilon$, we store the error and corresponding IK solutions in Line $8$ and $9$ respectively. Line $12$ finds the index $j$ corresponding to minimum error bound IK. In line $13$ we retrieve the robust-IK from the IK solution set.
\begin{algorithm}[!htb]
\caption{Steps to compute RobustIK\\ 
\textbf{Input}: $\bm{g}_d$, $c$, $\lambda$, $\epsilon \qquad$ \textbf{Output}: $\bm{\Theta}^*$}
\label{alg:SMB}
\begin{algorithmic}[1]
\State $\bm{p}_d \leftarrow \bm{g}_d[1:3, 4]$ and $\bm{q}_d \leftarrow \textbf{rotm\_2\_q}(\bm{g}_d[1:3, 1:3])$
\State $\bm{\Gamma} \leftarrow$ \textbf{IKsolver}$(\bm{g}_d)$ \qquad \text{where  } $\bm{\Gamma} \in \mathbb{R}^{M \times N}$
\For{\texttt{$(i \gets 1\ to\ M)$}}
    \State $\bar{\bm{\Theta}} \leftarrow \bm{\Gamma}[i]$
    \State $\mathbb{P} \leftarrow$ solve Eq~\eqref{eq: main_position_opti} using $\bm{p}_d, \bar{\bm{\Theta}}, c$
    \State $\mathbb{O} \leftarrow$ solve Eq~\eqref{eq:rotation_error_measure} using $\bm{q}_d, \bar{\bm{\Theta}}, c$
    \If{$\mathbb{P} + \lambda \mathbb{O} \leq \epsilon$}
	    \State $\mathbb{D}.\text{append}(\mathbb{P} + \lambda \mathbb{O})$
	    \State $sol.\text{append}(\bar{\bm{\Theta}})$
	\EndIf
\EndFor
\State $j \leftarrow \textbf{MinIndx}(\mathbb{D})$
\State $\bm{\Theta}^* \leftarrow sol[j, :]$
\Return $\bm{\Theta}^*$
\end{algorithmic}
\end{algorithm}
\noindent

\section{SIMULATIONS AND EXPERIMENTAL RESULTS}
\label{sec: numerical_ex}
\noindent
In this section we demonstrate the usefulness of our algorithm in the context of two applications: (i) Moving a parallel jaw gripper to a pre-grasp configuration to grasp a cubical object (ii) Moving a cylindrical peg to a pre-insertion configuration in a peg-in-hole assembly scenario. Here we present applications of robust-IK algorithm in two different situations. The application tasks are chosen such that in the first case only position error at the end-effector dictates success in the task and in the second case both position and orientation error at the end-effector space dictates success of the task. We present simulation and experimental results using $7$-DoF redundant Baxter research robot~\cite{baxterweb}.

\subsection{ROBUST IK FOR PRE-GRASP CONFIGUARTION}
\label{sec: bax_sim}
\noindent

\begin{figure}[!t]
	\centering
	\includegraphics[scale=0.60]{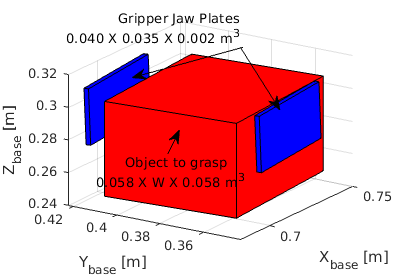}
    \caption{Simulation setup for placing end-effector in pre-grasp positioning task}	
	\label{fig: block_plates}
\end{figure}

In this example, a parallel jaw gripper has to move to a pre-grasp configuration such that it can grasp the object by closing it's grippers (see Fig\ref{fig: block_plates}). The gripper opening is $72$ mm. The size of the block is $58$ mm $\times$ $W$ mm $\times$ $58$ mm. The clearance between the gripper and the block is thus $(72 - W)/2$ mm. For our simulation study with Baxter robot, we vary $W$ from $58$ mm to $65$ mm to generate results for clearances between gripper and object varying from $7$ mm to $3.5$ mm. The size of the object and grippers are such that small position errors along $X_{base}$ and $Z_{base}$ directions, as well as orientation errors does not affect success of the task. However because of small clearance between the object and gripper opening, success of the task is sensitive to position error along $Y_{base}$. We call the pre-grasp configuration reaching task successful, if the minimum distance between one of the gripper faces and object is greater than the clearance. Thus, the error metric, $\epsilon$, is same as the distance between gripper frame and object frame along $Y_{base}$.

The joint error distribution for each joint is assumed to be a Gaussian, $\mathcal{N}(\mu = 0, \sigma=0.0045 \text{rad})$. The joint errors are assumed to be independent and drawn from the same distribution. The uncertainty set in the joint space is considered to be the set with $95\%$ probability mass, (i.e., $k=2$ in Equation~\eqref{eq:js_error_ball}). The choice of $\sigma$ is such that $95\%$ of the joint error is within $\pm 0.51$ degrees of the desired configuration. These error parameters are in accordance with Baxter robot~\cite{baxterweb}.

\begin{figure}[!t]
	\centering
	\includegraphics[scale=0.45]{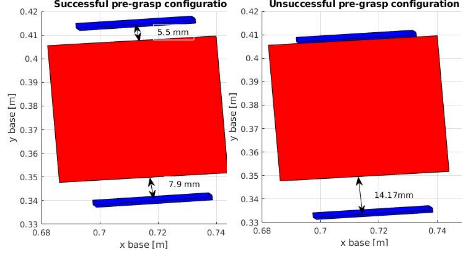}
	\caption{Successful and failed instances of positioning tasks by $\bm{\Theta}^*$ and $\bm{\Theta}^-$ respectively}
	\label{fig: collision_instance}
\end{figure}

\begin{figure}[!htb]
	\centering
	\includegraphics[scale=0.5]{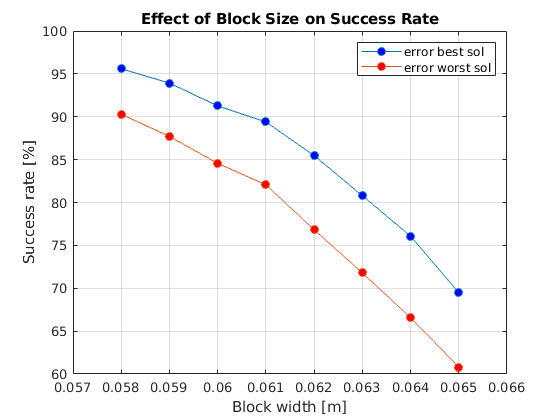}
	\caption{Success rate of best and worst solutions with varying block size while the gripper opening is fixed at $72$mm}
	\label{fig: block_plates_success}
\end{figure}

The desired end-effector pose is $\bm{p}_d = [0.71305, 0.3786, 0.300]$ and $\bm{q}_d = [0.0086, 0.9992, 0.0370, 0.0155]$.  Using Algorithm~\ref{alg:SMB} we found the best IK solution as $\bm{\Theta}^* = [0.0052, -0.1660, -2.0927, 1.1777, 1.6105, 2.0793, 2.6467]$rad. In Fig~\ref{fig: collision_instance} instances of successful and unsuccessful pre-grasp positioning of the end-effector is shown for best and worst IK solutions respectively. In Fig~\ref{fig: block_plates_success} we have plotted success-rates in accomplishing the pre-grasp positioning task by best IK solution $(\bm{\Theta}^*)$ along with the worst IK solution $(\bm{\Theta^}-)$ (IK associated to maximum error bound) while the clearance is varied by varying block width. Each point in the plot is generated by averaging over $100$ randomly chosen joint space configuration with the error drawn from $\mathcal{N}(0, 0.0045)$. Note that when block width is $w=58$mm and desired success-rate is $> 80\%$ it does not matter if $\bm{\Theta}^*$ or $\bm{\Theta}^-$ is executed because both have success rate more than $90\%$. However if block width is $w=63$mm only $\bm{\Theta}^*$ has success-rate $>80\%$, hence this solution should be executed instead of $\bm{\Theta}^-$. Further if block width is $w=65$mm none of the IK solution has desired success-rate indicating that the robot will like fail in the grasping task for the block width of $65$ mm.

\noindent

\begin{figure}[!]
	\centering
	\includegraphics[scale=0.3]{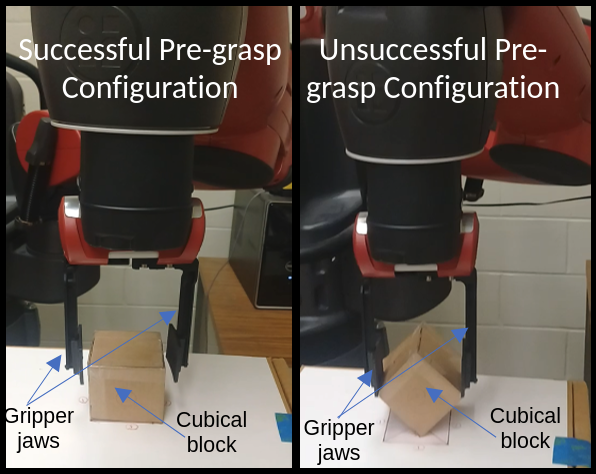}
	\caption{Instances of successful and failed pre-grasp positioning of end-effector with best and worst solutions}
	\label{fig: baxter_exp_success_failure}
\end{figure}

\noindent
\textbf{Experimental Result: }
We performed preliminary experiments with Baxter robot for the same task as in simulation with block width of $58$ mm and $\bm{\Theta}^*$ and $\bm{\Theta}^-$ computed as before. For $10$ different trials, $\bm{\Theta}^*$ and $\bm{\Theta}^-$ are executed from different initial end-effector configurations and outcomes are tabulated in Tab~\ref{tab: exp_success}. The Table shows that the best configuration was successful in $9$ out of $10$ cases, whereas the worst configuration was successful in $8$ out of $10$ cases. In Fig~\ref{fig: baxter_exp_success_failure} we have shown instances of successful and failed pre-grasp positioning of end-effector after executing $\bm{\Theta}^*$ and $\bm{\Theta}^-$ respectively. We considered the positioning as unsuccessful, if the gripper hit the object while reaching the end effector configuration. To make sure that the collision was due to the final gripper configuration and not because of the motion plan, we take care to place the gripper initially at a configuration so that the gripper can move straight down to reach the desired configuration. 


\begin{table}
\centering
\begin{tabular}{||c | c | c | c | c | c | c | c | c | c | c ||}
\hline
Trials & 1 & 2 & 3 & 4 & 5 & 6 & 7 & 8 & 9 & 10 \\ [0.5ex]
\hline
$\Theta_{rob}$ & 1 & 1 & 1 & 1 & 1 & 0 & 1 & 1 & 1 & 1 \\
$\Theta_{wst}$ & 0 & 1 & 0 & 1 & 1 & 0 & 1 & 1 & 0 & 1 \\
\hline
\end{tabular}
\caption{Successful and unsuccessful events for best and worst solutions with respect to each trial. Successful anf unsuccessful events are denoted as $1$ and $0$ respectively}
\label{tab: exp_success}
\end{table}

\subsection{ROBUST IK FOR PEG-IN-HOLE TYPE PROBLEMS}
\begin{figure}[!htb]
    \centering
    \includegraphics[scale=0.2]{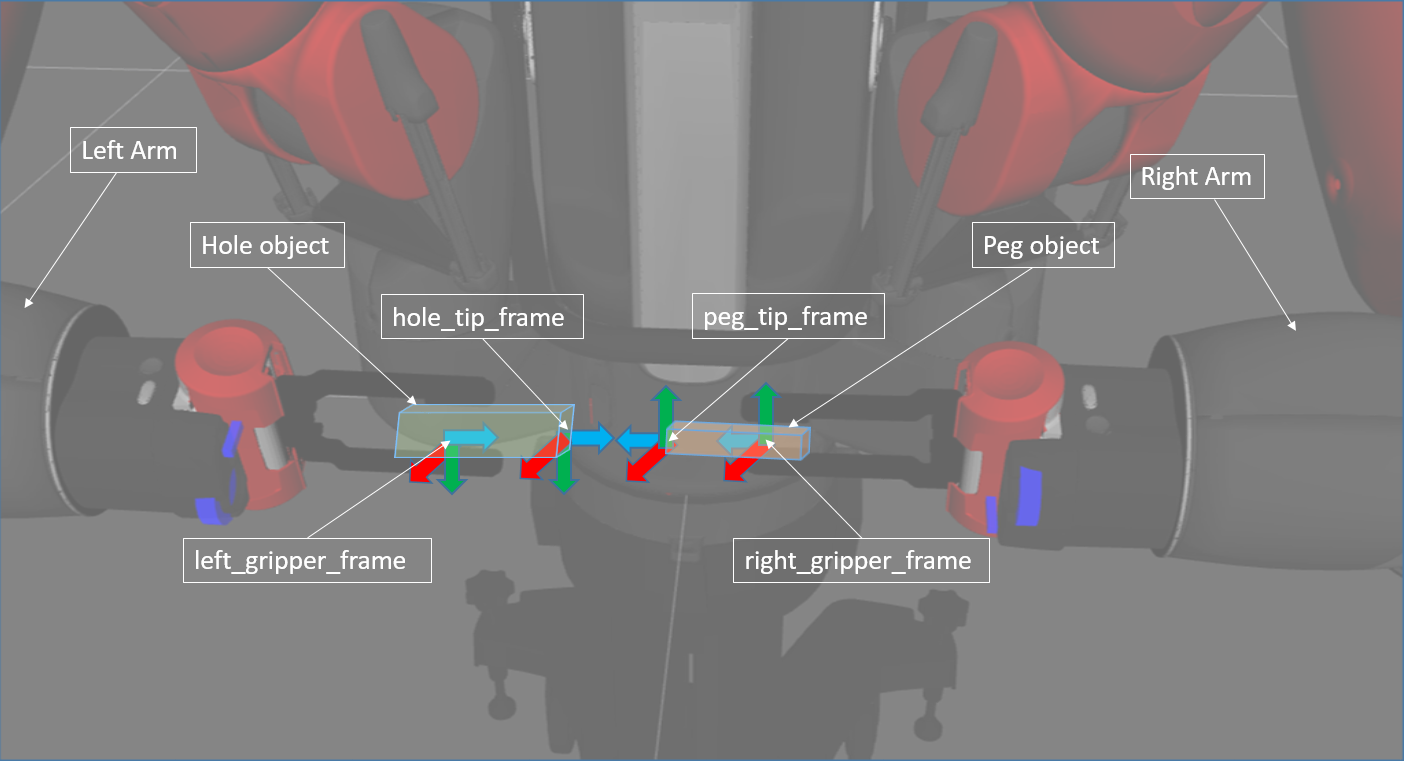}
    \caption{Simulation setup for peg-in-hole task. Hole is kept fixed by right arm while the left arm has to position peg\_tip frame within an error margin (based on available clearance) around the desired pre-insertion position. Frames of interests are also shown}
    \label{fig: pose_error_schematic}
\end{figure}

\begin{figure}[!htb]
    \centering
    \includegraphics[scale=0.4]{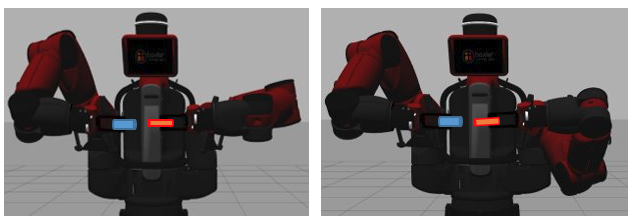}
    \caption{Instances of the pose of the peg after executing best(left) and worst(right) IK solutions for peg-in-hole task}
    \label{fig:baxter_pose_error}
\end{figure}

\noindent
In this example, we study the problem of peg-in-a-hole insertion (see Fig~\ref{fig:baxter_pose_error}). Our goal is to place a cylindrical peg in a pre-insertion configuration and then perform the assembly by moving the peg along the axis of the cylinder. Any pose error at the end-effector frame {i.e., } left\_gripper frame manifests as position error at the peg\_tip frame. Let $(\bm{p}_d,\bm{q}_d)$ is the desired and $(\bm{p},\bm{q})$ is the achieved configurations at peg\_tip frame. Then from position error $(e_{peg\_tip})$ at peg\_tip frame as in Eq~\eqref{eq: error_peg_tip} we can derive error metric to minimize in Eq~\eqref{eq: err_peg_assembly}.
\begin{eqnarray}
    \label{eq: error_peg_tip}
    e_{peg\_tip} & = & (\bm{p} - \bm{p}_d) + l_p  (\bm{R}_z - \bm{R}_{d_z})\\ \nonumber
    ||e_{peg\_tip}|| & = & ||(\bm{p} - \bm{p}_d) + l_p  (\bm{R}_z - \bm{R}_{d_z})||\\ \nonumber
    & \leq & ||\bm{p} - \bm{p}_d|| + l_p||\bm{R}_z - \bm{R}_{d_z}||\\
    \label{eq: err_peg_assembly}
    & = & ||\bm{p} - \bm{p}_d|| + l_p|\theta_z|
\end{eqnarray}
where $\bm{R}_z$ and $\bm{R}_{d_z}$ represents $z-$axes of realized and desired rotation matrices and $\theta_z$ is the included angle between them. The term $||\bm{p} - \bm{p}_d||$ is equivalent to $\mathbb{P}$ of Eq~\eqref{eq: opti_position_only}, $|\theta_z|$ is equivalent to $\mathbb{O}$ of Eq~\eqref{eq: opti_orientation_only} and $l_p$ the length of the peg extended from left\_gripper frame is the choice of $\lambda$.   

\begin{figure}[!htb]
    \centering
    \includegraphics[scale=0.5]{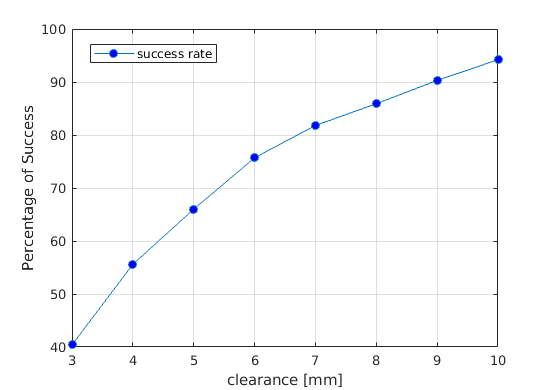}
    \caption{Increase in success rate with increased clearance between peg and hole for best IK solution}
    \label{fig: pose_error_best_sol}
\end{figure}

The error model in the joint space is same as that of the previous example. The length of the peg, $l_p=0.10$ m. The diameter of the hole, $d_{H} = 24$ mm, and the diameter of the peg $d_{P}$ varies from $4$ mm and $18$ mm. The clearance between the peg and the hole is $(d_{H} - d_{P})/2$ and it varies from $10$ mm and $3$ mm.
A pre-insertion configuration is considered successful if the error $e_{peg\_tip}$ is smaller than the clearance.
In the simulations, the desired $left\_gripper$ frame configuration  of the peg is $\bm{p}_d=[0.6165, 0.077, 0.4025]$m, $\bm{q}_d=[0.6839, 0.7174, 0.0799, -0.1064]$. Then using Algorithm\ref{alg:SMB} we find best IK solution as $\bm{\Theta}^*=[0.365997,\,-0.205692,\,-1.45802,\,1.66477,\,2.93037,\,-1.12361,$ $-0.142083]$. Figure~\ref{fig: pose_error_best_sol} shows the success rate of the best (robust) IK as a function of the clearance. Each data point is generated by averaging over $1000$ random runs with the joint space error drawn from $\mathcal{N}(0,0.0045)$. From the figure, we can see that for the peg diameter of $10$ mm,  more than $80\%$ of time the IK solution was able to place end-effector ({i.e., }left\_gripper frame) inside the allowable neighborhood of desired pose resulting in successful placement of the peg for assembly. This plot also shows that given the error tolerance, the robot can evaluate it's capability to successfully complete the job. For example, for a clearance of $3$ mm, the robot can evaluate that the assembly plan will likely fail most of the time. Figure~\ref{fig:baxter_pose_error} shows instances of placing end-effector to desired pre-insertion pose with best and worst IK solutions for $d_H = 10$ mm.


\section{CONCLUSION}
\label{sec: conclusion}
In this paper, we formulated and presented a method for solving the robust inverse kinematics problem. The performance of a robot in many tasks like grasping or assembly is dependent on the error of the robot in reaching the pre-grasp or pre-insertion configuration.
Because of inherent actuation errors in joint space, robots cannot achieve desired configurations in task space exactly. Furthermore, different inverse kinematics (IK) solutions map joint space error set to task space differently. Thus for a given task with a prescribed error tolerance, not all IK solutions will not be guaranteed to successfully execute the task. The robust IK solution gives the robots an IK solution that is robust to joint space errors. It also gives the robots a capability to evaluate the likelihood of succeeding in a particular task. 

We formalized the robust IK (more precisely it'sminimization version) problem as min-max type constraint optimization problem. We exploit the dependencies of the constraint variables and objective structure to reduce the main optimization problem into two smaller optimization problems. We showed that each of these two sub-problems can be posed as finding maximum eigenvalue problems. Finally using simulation and experimental results we show that computing robust-IK is indeed helpful in deciding feasibility of tasks knowing robot kinematic model and joint error information.

\textbf{Future Work:} In future we plan to see how self evaluation application of robust IK algorithm can be utilized when there are uncertainties both in actuation of the robot and pose of the block to grasp. We also plan to extend this research in propagating errors to task space of a mobile manipulator with actuation errors in its base and arm joints.

\textbf{ACKNOWLEDGEMENT}\\
\noindent
This work was supported in part by AFOSR award FA9550-15-1-0442.

\bibliographystyle{asmems4}
\bibliography{references}
\end{document}